# When AI Gets it Wrong: Reliability and Risk in AI-Assisted Medication Decision Systems

Khalid Adnan Alsayed
*Founder, Ducaltus | AI Assurance & Governance*
*BSc (Hons) Artificial Intelligence*
*School of Computing, Engineering & Digital Technologies*
Teesside University, UK
hello@ducaltus.com
s.khalid.adnan@gmail.com

**Abstract**—Artificial intelligence (AI) systems are increasingly integrated into healthcare and pharmacy workflows, supporting tasks such as medication recommendations, dosage determination, and drug interaction detection. While these systems often demonstrate strong performance under standard evaluation metrics, their reliability in real-world decision-making remains insufficiently understood. In high-risk domains such as medication management, even a single incorrect recommendation can result in severe patient harm.

This paper examines the reliability of AI-assisted medication systems by focusing on system failures and their potential clinical consequences. Rather than evaluating performance solely through aggregate metrics, this work shifts attention towards how errors occur and what happens when AI systems produce incorrect outputs. Through a series of controlled, simulated scenarios involving drug interactions and dosage decisions, we analyse different types of system failures, including missed interactions, incorrect risk flagging, and inappropriate dosage recommendations.

The findings highlight that AI errors in medication-related contexts can lead to adverse drug reactions, ineffective treatment, or delayed care, particularly when systems are used without sufficient human oversight. Furthermore, the paper discusses the risks of over-reliance on AI recommendations and the challenges posed by limited transparency in decision-making processes.

This work contributes a reliability-focused perspective on AI evaluation in healthcare, emphasising the importance of understanding failure behavior and real-world impact. It highlights the need to complement traditional performance metrics with risk-aware evaluation approaches, particularly in safety-critical domains such as pharmacy practice.

## 1. Introduction

Artificial intelligence (AI) has become increasingly integrated into healthcare systems, including pharmacy practice, where it is used to support tasks such as medication recommendations, dosage determination, and drug interaction detection [1], [2]. These systems are designed to enhance clinical decision-making, improve efficiency, and reduce human error. As a result, AI-assisted tools are being adopted in various stages of medication management, from prescribing to dispensing and monitoring.

Despite these advancements, the reliability of AI systems in real-world healthcare environments remains a critical concern. While many AI models demonstrate strong performance under standard evaluation metrics, such as accuracy and precision, these measures do not fully capture how systems behave when they encounter unexpected or complex scenarios [3]. In high-stakes domains such as pharmacy, even a single incorrect recommendation can lead to serious consequences, including adverse drug reactions, ineffective treatment, or patient harm.

A key issue lies in the growing reliance on AI systems without sufficient understanding of their limitations. Healthcare professionals may place a high degree of trust in AI-generated recommendations [4], particularly when systems appear to perform consistently well under evaluation. However, this trust can become problematic when systems fail in ways that are not immediately obvious, especially in cases where errors are not easily detectable or explainable. This creates a gap between perceived system performance and actual reliability in practice.

Furthermore, current approaches to evaluating AI systems often focus on aggregate performance metrics, which can obscure critical failure behaviors [5]. These evaluations typically do not account for the nature of errors or their real-world impact, particularly in safety-critical contexts such as medication decision-making. As a result, important questions remain unanswered: What happens when AI systems produce incorrect outputs? How do these errors affect clinical outcomes? And to what extent can such systems be safely relied upon in pharmacy practice?

This paper addresses these concerns by shifting the focus from performance to reliability in AI-assisted medication decision systems. It explores how system failures occur, examines their potential clinical consequences, and highlights the risk associated with over-reliance on AI in pharmacy contexts. By analysing controlled, simulated scenarios, this work aims to provide a clearer understanding of how errors in AI systems translate into real-world risks, and why reliability should be a central consideration in the evaluation and deployment of such technologies.

## 2. The Reliability Problem in AI-Assisted Medication Systems

Artificial intelligence systems used in healthcare and pharmacy are often evaluated based on their performance under controlled conditions. However, strong performance in testing environments does not necessarily translate to reliable behavior in real-world clinical settings. This discrepancy highlights a fundamental issue: AI systems may appear accurate and effective yet still fail in unpredictable or unsafe ways when deployed in practice [5], [6].

One of the key challenges is that AI systems do not always behave consistently across different inputs or contexts. Small variations in patient data, drug combinations, or clinical conditions can lead to significantly different outputs. In medication-related decision systems, such variability can result in inconsistent recommendations, which may not always be immediately detected by healthcare professionals [6].

Furthermore, many AI systems operate as black-box models, where the reasoning behind a given recommendation is not easily interpretable. This lack of transparency makes it difficult for practitioners to understand why a particular decision was made, reducing their ability to identify potential errors. As a result, incorrect recommendations may be accepted without sufficient scrutiny, particularly in time-constrained clinical environments [7].

Another important issue is that reliability is often overshadowed by performance-focused evaluations. Metrics such as accuracy, precision, or recall provide an overall indication of model performance but do not capture how systems behave when they fail. In safety-critical domains such as pharmacy, understanding failure is essential, as even rare errors can have severe consequences [3], [8].

This creates a gap between how AI systems are evaluated and how they are used in practice. While systems may meet performance benchmarks, their reliability in real-world decision-making scenarios remains uncertain. Addressing this gap requires a shift in focus towards analysing system behavior under failure conditions and understanding the risks associated with incorrect outputs in medication-related contexts.

## 3. Types of Failures in Medication Decision Systems

AI-assisted decision systems can fail in several distinct ways, each carrying different levels of clinical risk. Understanding these failure types is essential for evaluating system reliability, particularly in pharmacy contexts where incorrect recommendations can directly impact patient safety. Unlike general performance metrics, analysing failure types provides insight into how and why systems produce incorrect outputs and what the consequences of these errors may be [5], [6].

One of the most critical types of failure is the false negative, where a system fails to identify a genuine risk. In medication-related scenarios, this may occur when an AI system does not detect a harmful drug interaction or overlooks a contraindication. Such errors are particularly dangerous because they can lead to severe adverse drug reactions without any warning, giving both the system and the user a false sense of safety [8].

In contrast, false positives occur when a system incorrectly flags a safe situation as risky. For example, an AI system may identify a drug interaction where none exists, leading to unnecessary changes in treatment. While less immediately dangerous than false negatives, false positives can result in delayed care, reduced treatment effectiveness, and increased complexity in clinical decision-making [2].

Another important failure type involves incorrect dosage recommendations. AI systems may suggest dosages that are either too high or too low for a specific patient. Overestimation can lead to overdose and toxicity, while underestimation may result in ineffective treatment. Dosage-related errors are particularly concerning in pharmacy practice, as they require precise consideration of patient-specific factors such as age, weight, and medical history [2], [8].

These failure types highlight that not all errors are equal in their impact. Some errors may have minimal consequences, while others can be life-threatening. Therefore, evaluating AI systems purely based on aggregate performance metrics fails to capture the severity and distribution of these errors. A reliability-focused approach must consider both the type of error and its potential clinical outcome.

*Table I. Types of AI errors in medication decision systems*

| Error Type | Example Scenario | Potential Impact |
| --- | --- | --- |
| False Negative | Missed Drug Interaction | Severe Adverse Reaction |
| False Positive | Incorrect Interaction Flagged | Unnecessary Treatment Change |
| Wrong Dosage | Incorrect Dosage Recommendation | Overdose or Ineffective Care |

## 4. Clinical Risk and Pharmacy Impact

Artificial intelligence errors in medication decision systems are not merely technical issues; they translate directly into clinical risks that can affect patient safety and treatment outcomes. In pharmacy practice, decisions related to drug selection, dosage, and interaction management require a high degree of precision. When AI systems produce incorrect outputs in these contexts, the consequences can be immediate and severe, particularly if errors are not identified before implementation [2], [8].

One of the most critical risks arises from undetected drug interactions. If an AI system fails to identify a harmful interaction between medications, patients may experience adverse drug reactions ranging from mild side effects to life-threatening conditions. These risks are especially significant in patients with complex medication regimens, where multiple drugs are prescribed simultaneously and interactions are more difficult to detect manually [8].

Incorrect dosage recommendations represent another major area of concern. Dosage decisions must account from individual patient factors such as age, weight, and comorbidities. AI systems that do not fully capture these variables or that misinterpret input data may recommend inappropriate dosages. Overdosing can lead to toxicity and severe complications, while underdosing may result in ineffective treatment and disease progression [2].

In addition, false positive errors can disrupt clinical workflow and lead to unnecessary interventions. When AI systems incorrectly flag safe medications as dangerous, healthcare professionals may alter treatment plans unnecessarily, causing delays or reducing treatment effectiveness. While these errors may not always result in direct harm, they contribute to inefficiencies and can undermine trust in both the system and clinical process [4].

These risks are further amplified by the increasing reliance on AI systems in healthcare environments. As clinicians and pharmacists begin to depend more heavily on automated recommendations, the likelihood of overlooking system errors increases. This over-reliance can reduce critical evaluation of AI outputs, particularly when systems are perceived as highly accurate or authoritative [4].

Overall, the clinical impact of AI failure in medication decision systems highlights the importance of moving beyond performance-based evaluation. Understanding how errors manifest in real-world scenarios, and the severity of their consequences, is essential for ensuring safe and effective integration of AI into pharmacy practice.

## 5. Experimental Design and Simulated Evaluation

To better understand how AI system failures translate into real-world risks, this study employs a controlled, simulated evaluation of medication decision scenarios. Rather than relying on real patient data, which introduces ethical and regulatory challenges, a set of representative cases was constructed to model common pharmacy-related decision situations. This approach allows for focused examination of system behavior under failure conditions while maintaining clarity and reproducibility.

A total of ten medication scenarios were defined, each representing a typical decision-making situation encountered in pharmacy practice. These scenarios included cases involving drug interactions, dosage recommendations, and safe medication combinations. For each scenario, a ground truth outcome was established based on standard clinical knowledge, indicating whether the medication combination or dosage was safe or potentially harmful [8], [9].

Simulated AI outputs were then assigned to each case to represent both correct and incorrect system behavior. These outputs were intentionally designed to include different types of errors, such as missed drug interactions (false negatives), incorrect risk alerts (false positives) and inappropriate dosage recommendations. This allowed for a structured analysis of how different failure types manifest in medication-related contexts and what their potential consequences may be.

The results of this simulated evaluation highlight that even a small number of incorrect outputs can lead to significant clinical risks. For example, in scenarios where drug interactions were not detected, the potential for severe adverse reactions was observed. Similarly, incorrect dosage recommendations demonstrated how system errors could result in either toxicity or ineffective treatment. These findings reinforce the importance of examining failure behavior rather than relying solely on overall system performance [9], [10].

To support this analysis, the simulated cases and their corresponding outcomes are summarized in Table 2. The table presents each scenario, the AI system output, and the actual clinical assessment, and the resulting error classification. This structured representation provides a clear overview of how different types of failures occur and why they are critical in medication decision systems.

The simulated evaluation showed that false negatives were the most frequent error type (n = 3), followed by wrong dosage errors (n = 2), while false positives were less common (n = 1). This highlights the increased risk associated with undetected harmful interactions in AI-assisted medication systems.

This distribution of error types observed in the simulated evaluation is illustrated in Figure 1.

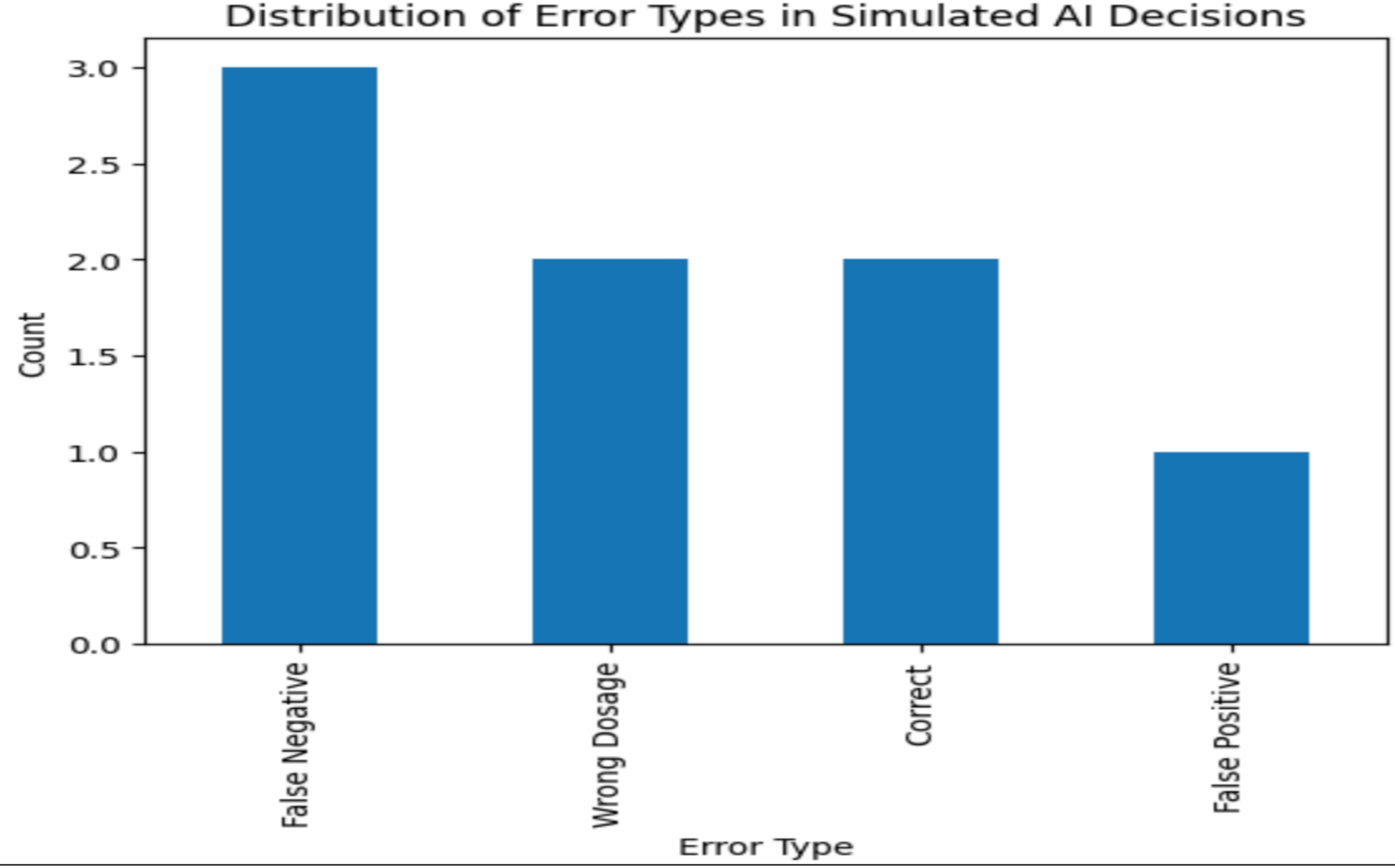

*Figure 1. Distribution of error types in simulated AI-assisted medication decision scenarios.*

*Table II. Simulated AI decision outcomes in medication scenarios*

| Case | Scenario | AI Output | Actual Outcome | Error Type |
|---|---|---|---|---|
| 1 | Drug A + Drug B Interaction | Safe | Dangerous | False Negative |
| 2 | Drug C + Drug E Combination | Dangerous | Safe | False Positive |
| 3 | High Dosage Drug E | Safe Dose | Overdose | Wrong Dosage |
| 4 | Drug F + Drug G Interaction | Safe | Dangerous | False Negative |
| 5 | Low Dosage Drug H | High Dose | Safe Dose | Wrong Dosage |
| 6 | Drug I + Drug J Interaction | Dangerous | Dangerous | Correct |
| 7 | Standard Dosage Drug K | Safe Dose | Safe Dose | Correct |
| 8 | Drug L + Drug M Interaction | Safe | Dangerous | False Negative |

## 6. Discussion

The results of the simulated evaluation highlight a critical issue in AI-assisted medication decision systems: system errors are not only possible but can have significantly different levels of clinical impact depending on their type. While overall system performance may appear acceptable, the distribution and nature of errors reveal underlying risks that are not captured by aggregate evaluation metrics. This reinforces the need to move beyond performance-focused assessment towards a more reliability-oriented perspective [3], [5], [11].

One of the most significant findings is the prevalence of false negatives within the simulated scenarios. These errors, where harmful drug interactions or risks are not identified, represent the most dangerous forms of system failure in medication-related contexts. Unlike false positives, which may lead to cautions or unnecessary changes in treatment, false negatives provide a false sense of safety and can result in direct patient harm. This asymmetry in risk has been highlighted in clinical safety research, where missed errors are often associated with more severe outcomes than overly cautious decisions [8], [10], [12].

In addition, the presence of dosage-related errors further illustrates the complexity of AI-assisted decision-making in pharmacy. Dosage recommendations require consideration of multiple patient-specific variables, and even small inaccuracies can lead to significant clinical consequences. Prior work in clinical pharmacology emphasises that medication dosing errors remain a major contributor to preventable harm, particularly in systems that rely on automated support tools [2], [13].

Another key issue is the potential for over-reliance on AI systems by healthcare professionals. As AI tools become more integrated into clinical workflows, there is an increased risk that users may accept recommendations without sufficient verification. This is particularly problematic when systems lack transparency or explainability, making it difficult for practitioners to assess the validity of outputs. Studies in human-AI interactions have shown that automation bias and overconfidence in decision-support systems can reduce oversight and increase the likelihood of undetected errors [4], [7], [14].

Furthermore, the findings emphasise that reliability is not solely a technical property, but a practical concern that emerges from the interaction between AI systems and human users. Even a highly accurate system can be unsafe if its errors are not well understood or if its outputs are misinterpreted. Recent research in trustworthy AI highlights that system reliability must be evaluated in context, considering usability, interpretability, and real-world deployment conditions [6], [11], [15].

Overall, this study demonstrates that focusing on reliability and failure behavior provides a more meaningful understanding of AI performance in medication decision systems. By examining how errors occur and what their consequences are, it becomes possible to identify risks that would otherwise remain hidden under traditional evaluation approaches. This shift in perspective is essential for ensuring the safe and effective integration of AI into pharmacy practice.

## 7. Recommendations

The findings of this study highlight the need for a more cautious and structured approach to the deployment of AI-assisted medication decision systems. While these systems offer significant potential to improve efficiency and support clinical workflows, their integration into pharmacy practice must be accompanied by safeguards that address reliability and risk.

First, AI systems used in medication-related decision-making should not be relied upon as standalone tools. Human oversight must remain a central component of the decision-making process, particularly in high-risk scenarios involving drug interactions and dosage recommendations. Healthcare professionals should be encouraged to critically evaluate AI outputs rather than accept them as authoritative, ensuring that errors can be identified before they impact patient care [4], [14].

Second, evaluation practices for AI systems should be expanded beyond traditional performance metrics. As demonstrated in this study, aggregate measures such as accuracy fail to capture the nature and severity of system errors. Instead, evaluation frameworks should incorporate analysis of failure types, error distributions, and clinical impact. This includes placing greater emphasis on high-risk errors, such as false negatives, which may lead to severe patient harm if undetected [3], [11].

Third, improving transparency and interpretability in AI systems is essential for enhancing trust and usability. Systems that provide clear reasoning or explanations for their recommendations allow healthcare professionals to better understand and validate outputs. This reduces the likelihood of blind trust and supports more informed decision-making in clinical environments [7], [15].

Finally, future development of AI systems in pharmacy should prioritise safety-oriented design. This includes incorporating mechanisms for uncertainty estimation, alert prioritization, and fail-safe behaviors when confidence in predictions is low. By focusing on reliability and risk mitigation, AI systems can be better aligned with the safety requirements of healthcare practice [5], [6].

## 8. Conclusion

Artificial intelligence systems are becoming increasingly integrated into pharmacy and healthcare workflows, offering valuable support in medication-related decision-making. However, this study demonstrates that strong performance under standard evaluation metrics does not necessarily translate to reliable or safe behavior in real-world settings. In high-risk domains such as medication management, even a small number of errors can lead to significant clinical consequences.

Through a controlled, simulated evaluation, this paper examined how different types of system failures, particularly false negatives, false positives, and dosage-related errors can manifest in AI-assisted medication decision systems. The findings show that not all errors carry equal risk, and that certain failure types, such as undetected harmful interactions, can have severe implications for patient safety. These results highlight the limitations of performance-focused evaluation approaches, which often fail to capture the nature and impact of system errors.

This work contributes a reliability-focused perspective on the evaluation of AI systems in healthcare. By shifting attention towards failure behavior and real-world risk, it provides a more meaningful framework for understanding how AI systems should be assessed in safety-critical contexts. The inclusion of simulated scenarios and structured analysis further demonstrates how even simple evaluations can reveal important insight into system behavior.

Ultimately, ensuring the safe and effective use of AI in pharmacy requires a balanced approach that combines technological advancement with careful consideration of reliability, transparency, and human oversight. Future work should continue to explore methods for evaluating and mitigating risk in AI-assisted decision systems, with a particular focus on aligning system performance with the practical realities of clinical use.